\documentclass[twocolumn, longbibliography]{revtex4-2}

\newcommand*{\Title}{Advancing Stochastic 3-SAT Solvers by Dissipating Oversatisfied Constraints}
\newcommand*{\Author}{Joachim Schwardt}

\usepackage{color}
\usepackage{graphicx}
\usepackage{gensymb}
\usepackage{amsfonts}
\usepackage{amssymb}
\usepackage{amsmath}
\usepackage{amsthm}
\usepackage{array}
\usepackage{esint}
\usepackage{bm, bbm} 
\usepackage{booktabs}
\usepackage{multirow}
\usepackage{pgffor}
\usepackage{svg}
\usepackage{hyperref}
\usepackage{csquotes}
\usepackage{mathtools}
\usepackage[all]{nowidow}
\usepackage{placeins}
\usepackage{microtype}

\usepackage{listings}
\usepackage{algpseudocode}
\usepackage{algorithm}

\makeatletter
\newcommand{\citecomment}[2][]{\citealp{#2}#1\citevar}
\newcommand{\citeone}[1]{\citecomment{#1}}
\newcommand{\citetwo}[2][]{\citecomment[,~#1]{#2}}
\newcommand{\citevar}{\@ifnextchar\bgroup{;~\citeone}{\@ifnextchar[{;~\citetwo}{]}}}
\newcommand{\citefirst}{\@ifnextchar\bgroup{\citeone}{\@ifnextchar[{\citetwo}{]}}}

\makeatother

\usepackage[english]{babel}
\usepackage[T1]{fontenc}
\usepackage{microtype}
\usepackage[english,capitalise]{cleveref}

\usepackage[acronym]{glossaries}

\makeatletter
\let\oldtheequation\theequation
\renewcommand\tagform@[1]{\maketag@@@{\ignorespaces#1\unskip\@@italiccorr}}
\renewcommand\theequation{(\oldtheequation)}
\makeatother

\newcommand{\algoref}[1]{\hyperref[#1]{Algorithm~\ref*{#1}}}
\newcommand{\appref}[1]{\hyperref[#1]{Appendix~\ref*{#1}}}

\makeatletter
\newcommand*{\glsplainhyperlink}[2]{%
    \begingroup%
      \hypersetup{hidelinks}%
      \hyperlink{#1}{#2}%
    \endgroup%
}
\let\@glslink\glsplainhyperlink
\makeatother

\hypersetup{
    colorlinks,
    breaklinks,
    citecolor=blue,
    linkcolor=blue,
    urlcolor=blue,
    pdfauthor={\Author},
    pdftitle={\Title}
}

\newcommand{\walksat}{{WalkSAT}}
\newcommand{\probsat}{{ProbSAT}}
\newcommand{\yalsat}{{YalSAT}}
\newcommand{\kissat}{{Kissat}}
\newcommand{\rowsat}{{DOCSAT}}
\newcommand{\row}{{\text{doc}}}

\newcommand{\myparagraph}[1]{{\bfseries{#1}.}}

\newacronym{sls}{SLS}{stochastic local search}
\newacronym{tlc}{TLC}{true literal count}
\newacronym{cnf}{CNF}{conjunctive normal form}
\newacronym{tl}{TL}{true literal}
\newacronym{fl}{FL}{false literal}
\newacronym{doc}{DOC}{dissipating oversatisfied constraints}

\newcommand{\jcb}[1]{\bgroup\color{orange} JCB: #1\egroup}
\newcommand{\jsc}[1]{\bgroup\color{purple!75!blue} JS: #1\egroup}
\newcommand{\TODO}[1]{\bgroup\color{red!60!black}TODO: #1\egroup}

\begin{document}

\title{\Title}
\author{\Author$^{1,2}$}
\email{jschwardt@pks.mpg.de} 
\author{Jan Carl Budich$^{2,1}$}

\affiliation{$^1$Max Planck Institute for the Physics of Complex Systems, N\"{o}thnitzer Str.~38, 01187 Dresden, Germany}
\affiliation{$^2$Institute of Theoretical Physics, Technische Universit\"{a}t Dresden and W\"{u}rzburg-Dresden Cluster of Excellence ct.qmat, 01062 Dresden, Germany}

\date{\today}

\begin{abstract}
We introduce and benchmark a stochastic local search heuristic for the NP-complete satisfiability problem 3-SAT that drastically outperforms existing solvers in the notoriously difficult realm of critically hard instances. 
Our construction is based on the crucial observation that well established previous approaches such as \walksat{} are prone to get stuck in local minima that are distinguished from true solutions by a larger number of oversatisfied combinatorial constraints. 
To address this issue, the proposed algorithm, coined \rowsat{}, dissipates oversatisfied constraints (DOC), i.e. reduces their unfavorable abundance so as to render them critical. 
We analyze and benchmark our algorithm on a randomly generated sample of hard but satisfiable 3-SAT instances with varying problem sizes up to $N = 15000$. 
Quite remarkably, we find that \rowsat{} outperforms both \walksat{} and other well known algorithms including the complete solver \kissat, even when comparing its ability to solve the hardest quintile of the sample to the average performance of its competitors. 
The essence of \rowsat{} may be seen as a way of harnessing statistical structure beyond the primary cost function of a combinatorial problem to avoid or escape local minima traps in stochastic local search, which opens avenues for generalization to other optimization problems.
\glsresetall
\end{abstract}

\maketitle

\section{Introduction}\label{sec:intro}
Hard combinatorial problems such as the NP-complete satisfiability problem 3-SAT \cite{sat.NP_complete.Cook_1971,np_complete.reducibility.combinatoric_problems.Karp_1972,np_complete.guide.computers_intractibility.Garey_1990} are of ubiquitous importance, both from a scientific perspective and for a wide range of applications from integrated circuit design \cite{sat.circuits.Nam_1999,sat.circuits.FPGA_islands.Mukherjee_2015}, model checking \cite{sat.model_checking.survey.Prasad_2005,sat.model_checking.verification.Gupta_2006} and logistics \cite{sat.timetables.Matos_2021,sat.routing.biochips.Yuh_2011,sat.routing.system_on_a_chip.Zhukov_2020} via frustrated magnetism \cite{np_complete.comp_complexity.ising_spinglass.Barahona_1982,np_complete.ising_formulations.lucas_2014} to protein folding \cite{sat.protein.optimization.Allouche_2014,sat.protein.Ollikainen_2009} and AI development \cite{ai.prob_reasoning.maxsat.Park_2002,ai.constraint_processing.Dechter_2003,ai.maxsat.maxsolver.Xing_2005,ai.MMP.MarquesSilva_2013}. 
Although a full solution to such universal problems remains elusive and is not even expected to become viable with the advent of quantum computers \cite{qc.cc.Preskill_2021,qc.democrit.Aaronson_2013}, their gravitas has motivated intense research efforts for decades.
Besides conceptual advances in complexity theory \cite{qc.democrit.Aaronson_2013}, impressive progress towards unlocking more and more complex 3-SAT problem instances has been made in at least two directions \cite{sat.model_checking.survey.Prasad_2005,sat.practical.review.complete.Kullmann_2008,sat.handbook.Biere_2009,sat.preprocessing.Een_2005,sat.CDCL.solvers.Silva_2009}.
First, the practical performance of so-called complete solvers proven to eventually master any instance, albeit at the expense of exponentially long worst-case runtime, has tremendously improved \cite{sat.revolution.Fichte_2023,sat.review.advances.Alouneh_2019}.
Second, stochastic solvers such as \walksat{} that heuristically approach the problem by sampling techniques frequently solve even hard instances up to quite remarkable problem sizes \cite{SLS.walksat.review.Hoos_2000,SLS.review.overview.Hoos_2015,SLS.WalkSAT.random_3sat.Fu_2020}.
Further improving on this latter class of stochastic computational tools is the main objective of our present article.

\begin{figure}[htp!]
    \centering
    \includegraphics{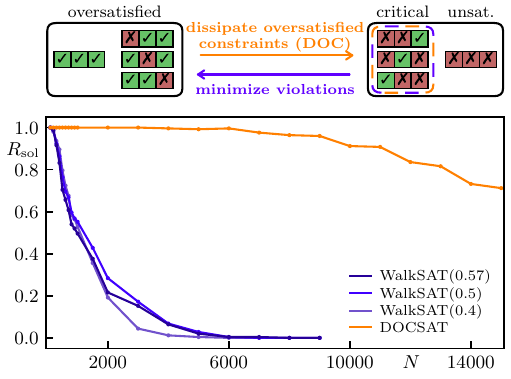}
    \caption{Upper panel: Illustration of two competing processes based on the 8 logical types of 3-SAT constraints.
    Minimizing violations, i.e. unsatisfied constraints, typically leads to excessive oversatisfied ones, which is counteracted by \acrshort{doc}.
    The desirable critical clauses are highlighted as a compromise of both processes.
    Lower panel: Fraction $R_\text{sol}$ of solved 3-SAT instances as a function of the number of variables $N$ for \walksat{} \cite{sls.walksat.Selman_1994} and \rowsat{} (cf.~\algoref{alg:rowsat}).
    At every $N$, 250 instances at critical clause density $\alpha=\alpha_\text{crit}=4.27$ are generated using the Weigt-protocol \cite{opt_alg_in_physics.chapter7.Weigt_2004}.
    The solvers are run $10^3$ times with $300N$ iterations per trial.
    The various walk probabilities $p_\text{walk}$ for \walksat{} are displayed in the legend, while $p_\text{walk}=0.4$ and $r_\row=0.15$ are fixed for \rowsat{}.
    }
    \label{fig:sol_ratio_wsat}
\end{figure}

Below, we introduce and benchmark a stochastic 3-SAT solver coined \rowsat{} (cf.~\algoref{alg:rowsat}) which provides a major advance in solving critically hard instances as compared to existing algorithms (see \autoref{fig:sol_ratio_wsat}). 
The underlying heuristic is based on the crucial insight that the landscape of local and global minima has additional structure beyond the natural 3-SAT cost function (energy) that simply counts unsatisfied constraints.
Specifically, we find that local minima in which \walksat{} is prone to get trapped for hard instances are distinguished from global minima (true solutions) by containing significantly more oversatisfied constraints (see \autoref{fig:wsat_tlc_stats}), i.e. combinatorial expressions that would remain logically true upon changing one of its correct Boolean variables (cf.~\autoref{fig:sol_ratio_wsat} upper panel).
Building on the celebrated \walksat{} heuristic \cite{sls.walksat.Selman_1994}, our \rowsat{} solver (see \autoref{fig:sol_ratio_wsat} upper panel and \autoref{fig:illustration} for illustration) exploits this structure by \gls{doc}, thus rendering more constraints critically satisfied so as to escape the aforementioned local minima towards true solutions of hard instances (cf.~\autoref{fig:hists_rowsat_wsat_f1000v0}).
Considering the hardest quintile from our sample of Weigt-type benchmark instances \cite{opt_alg_in_physics.chapter7.Weigt_2004} in the critical regime, we demonstrate that \rowsat{} performs stronger even on this unfavorable selection than \walksat{} does on the entire sample (see \autoref{fig:p_avg_wsat}). 
Finally, we verify with benchmark data that the \rowsat{} heuristic indeed generates critical clauses at the expense of oversatisified clauses at a higher rate than \walksat{}. 

\section{The structure behind low-energy caveats of \walksat{}}\label{sec:tlc_stats}
The motivation for our \rowsat{} solver (cf.~\autoref{sec:rowsat}) is based on new insights regarding the landscape of local minima that limit the performance of existing stochastic algorithms such as \walksat{}, with particular reference to hard critical 3-SAT instances generated by the Weigt-protocol \cite{opt_alg_in_physics.chapter7.Weigt_2004}.
To establish notation, we start by briefly discussing the key ingredients of our analysis, i.e. the 3-SAT problem and the \walksat{} heuristic.

A 3-SAT problem consists of $N$ Boolean  variables $\textbf{x}=(x_1,\dots,x_N)$ constrained by $M = \alpha N$ clauses, with the \emph{clause density} $\alpha$. 
Every clause $C_m$ consists of three literals, $C_m=l_{m,1}\lor l_{m,2} \lor l_{m,3}$, that can be either a variable (\emph{positive} literal, e.g. $l_{m,1}=x_{m_1}$) or its negation (\emph{negative} literal, e.g. $l_{m,1}=\bar{x}_{m_1}$). 
An assignment (or bitstring) $\textbf{x}$ solves the problem if all the clauses $\textbf{C}=(C_1,\dots,C_M)$ simultaneously evaluate to \emph{true} under it.
Formally, a 3-SAT problem is a Boolean formula in \gls{cnf}, i.e. $C_1 \land \ldots \land C_M$. 
The cost function, here called energy $E(\textbf{x})$, simply counts the number of violated constraints, such that solutions are distinguished by $E=0$. 
As an example, the clause $C_1=x_1\lor \bar{x}_3 \lor \bar{x}_7$ is satisfied by 7 out of the $2^3=8$ possible assignments to the involved triple of variables. 
Note that under insertion of an assignment, say $x_1=1$, $x_3=0$ and $x_7=1$, the individual literals of the clause $C_1$ become either \emph{true} ($x_1=1$, $\bar{x}_3=1$) or \emph{false} ($\bar{x}_7=0$). 
For a satisfying assignment $\textbf{x}$, every clause must contain at least one true literal. 
We refer to clauses with more than one true literal as \emph{oversatisfied}, as opposed to critical clauses (one true literal) and unsatisfied clauses (no true literal), respectively (cf.~\autoref{fig:sol_ratio_wsat} upper panel). 
Promoting critical clauses at the expense of oversatisfied clauses is at the heart of our present work. 

\Gls{sls} solvers approach a 3-SAT problem via a focused search starting from a random initial state $\textbf{x}$ \cite{sls.walksat.Selman_1994,SLS.walksat.review.Hoos_2000,SLS.review.overview.Hoos_2015,SLS.WalkSAT.random_3sat.Fu_2020}.
Focused means that every iteration of the algorithm starts by (randomly) selecting one of the unsatisfied clauses.
Guided by a heuristic, one then flips a variable associated with a literal in this clause, satisfying it in the process.
This is iterated for a fixed number of flips, and then the search can be restarted from a new random state.
The essence of such \gls{sls} algorithms then lies in the heuristic for selecting variables and clauses.

In the \gls{sls} heuristic \walksat{}, one keeps track of the breakcount $b$ for every variable, which is the number of clauses that would be broken by flipping that variable.
Then, with a certain probability $(1-p_\text{walk})$, one performs a greedy step by picking a variable with minimal breakcount in a randomly chosen unsatisfied clause.
With the remaining \emph{walk probability} $p_\text{walk}$, one instead mimics a random walk by flipping a variable at random within the aforementioned unsatisfied clause; this noise provides the major advantage over the GSAT heuristic \cite{sls.walksat.Selman_1994}.
Finally, if a breakcount is zero, that variable is always selected (i.e. ``free'' moves are preferred).
Note that \walksat{} can be recovered from our \rowsat{} heuristic (cf.~\algoref{alg:rowsat}, see \autoref{fig:illustration} for an illustration) by setting the parameter $r_\row=0$ in \autoref{eqn:doc_score}.
A \walksat{} implementation can be found in \cite{walksat}, which is the version we adapted for our benchmarks.

\myparagraph{True literal counting statistics}
We are now ready to reveal the crucial structure of low-energy minima that motivates our \rowsat{} heuristic. 
Using the Weigt-protocol \cite{opt_alg_in_physics.chapter7.Weigt_2004}, we generate 250 problem instances with $N=200$ variables with critical clause density $\alpha_\text{crit}=4.27$, which corresponds to the hardest regime of 3-SAT at a given $N$ \cite{sat.hard_regime.Cheeseman_1991,sat.hard_regime.distributions.Mitchell_1992,opt_alg_in_physics.chapter7.Weigt_2004}.
For each instance, we run \walksat{} for $10^4$ trials with $10^4$ flips per trial with walk probability $p_\text{walk} = 0.5$.
To study the low-energy landscape, we record the energy $E$ and \gls{tlc}, i.e. the total number of true literals, of every assignment occurring during the search. 
\begin{figure}[!htp]
    \centering
    \includegraphics{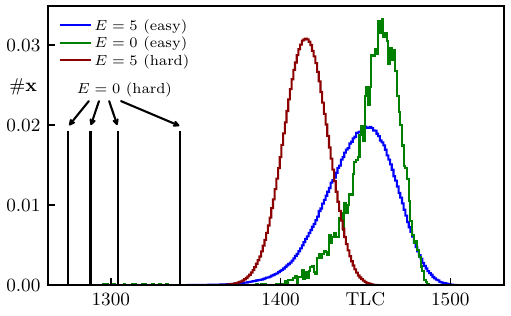}
    \caption{Density of configurations $\#\textbf{x}$ at a fixed energy $E$ as a function of \gls{tlc} for $10^4$ runs of \walksat{} with $10^4$ flips per run and $p_\text{walk} = 0.5$.
    Out of 250 instances with $N=200$, those with success probability $p > 90\%$ (easy) and $p < 1\%$ (hard) are compared.
    The black vertical stilts correspond to individual solutions for the hardest (for \walksat{}) four instances.
    For easy instances, differences between the $E=5$ bulk states and solutions ($E=0$) are small, while for hard instances the distribution at $E=5$ is largely disjoint in \gls{tlc} from the true solutions. 
    This indicates that the regime of low $E$ and large \gls{tlc} constitutes a vast landscape of local minima for \walksat{}.}
    \label{fig:wsat_tlc_stats}
\end{figure}

Generally speaking, \walksat{} shows strong variability in the success probability $p=\frac{\text{successful trials}}{\text{total trials}}$, and we refer to the extremes as (comparably) \emph{hard} ($p<1\%$) and \emph{easy} ($p>90\%$) instances, thus yielding 29 hard and 26 easy instances. 
Remarkably, as shown in \autoref{fig:wsat_tlc_stats}, we observe that the discrepancy between $\text{TLC}(E=0)$ and $\text{TLC}(E>0)$ is clearly correlated with $p$: For easy problems, the distribution of solution states ($E=0$) has \glspl{tlc} similar to that of the low-energy sector (exemplified by $E=5$). 
By contrast, for the hard problems, all of the (few) solutions found have significantly lower \gls{tlc}. This quantitatively substantiates the quite natural picture that solutions to hard instances should be fragile in the sense of combining many critical constraints. 
In summary, we conclude that \walksat{} is prone to ending up in a regime of high \gls{tlc}, a desert of low-energy minima that is far from a true solution. 

\section{The Dissipating oversatisfied constraints heuristic}\label{sec:rowsat}
Having identified a high \gls{tlc} as an obstruction to \walksat{}, we now introduce a new heuristic aimed at addressing this issue.
To start with, note that finding a state with minimal or maximal number of true literals is easy, because the problem factorizes into one for each variable:
to determine whether say $x_3$ should be set to true or false in the interest of \gls{tlc}, one merely needs to count the number of positive and negative occurrences of $x_3$ in the clauses, which can be stored in $N$-vectors $\textbf{p}$ and $\textbf{n}$ respectively.
A positive difference $(\textbf{p}-\textbf{n})_3$ then implies that $x_3=1$ in the assignment with maximal \gls{tlc} and $x_3=0$ in the one with minimal \gls{tlc}.

\begin{figure}[htp!]
    \centering
    \includegraphics{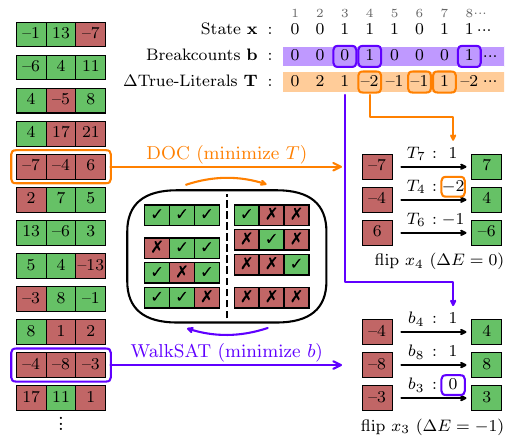}
    \caption{Illustration of the variable selection heuristic in \rowsat{} (see \algoref{alg:rowsat}) for a state $\textbf{x}$ and an exemplary 3-SAT instance, where a literal associated with variable $x_k$ ($\bar x_k$) is denoted by $k$ ($-k$).
    The number of critical clauses broken by flipping a variable $x_k$ is represented by $b_k$, and the change in the \gls{tlc} by $T_k$.
    The greedy step in \walksat{} flips the variable with minimal $b$.
    Instead \acrshort{doc} amounts to introducing the minimization of $T$, which leaves the fewest number of true literals after the flip.
    \rowsat{} uses a weighted compromise of both incentives by minimizing a weighted score (see \autoref{eqn:doc_score}).
    The central schematic illustrates the average effect of the individual processes on the true literal patterns: \walksat{} avoids breaking clauses, on average leading to more oversatisfied clauses, while \acrshort{doc} has the opposite effect (cf.~\autoref{fig:sol_ratio_wsat} upper panel).}
    \label{fig:illustration}
\end{figure}

Our goal is to reduce the \gls{tlc} by \acrfull{doc}. 
Instead of minimizing the breakcount $b$ as in \walksat{}, we can define a pure \acrshort{doc}-heuristic that minimizes the \gls{tlc}, leaving as few true literals as possible after a flip.
For a literal $l$ associated with the $k$-th Boolean variable, the change in the \gls{tlc} for flipping $x_k$ is given by
\begin{align}
    T=(\textbf{p}-\textbf{n})_{k}\cdot (2x_{k}-1). 
\end{align}
In \autoref{fig:illustration}, we illustrate the individual optimization with respect to $b$ and $T$ for a concrete example. Both figures of merit have their advantages and caveats. 
$b$ tends to reduce energy but accumulates oversatisfied constraints driving \walksat{} into the aforementioned local-minima desert. 
$T$ is designed to reduce oversatisfied constraints at the expense of occasionally increasing energy, thus the terminology \gls{doc}. 
As a trade-off aimed at combining the benefits of both, we introduce a total score  
\begin{align}
    s &= b + r_\row\cdot T, \label{eqn:doc_score}
\end{align}
where $r_\row$ is the relative weight of the \gls{doc}-term.
This implements a pull toward lower \gls{tlc} because the new term can act as a tiebreaker for identical breakcounts even for small $r_\row$, thus favoring critical clauses. 
The resulting pseudocode for selecting a variable to be flipped in a randomly chosen unsatisfied clause is given in \algoref{alg:rowsat}.
\begin{algorithm}[H]
\caption{Variable selection in \rowsat{}}\label{alg:rowsat}
\begin{algorithmic}[1]
\Require $\text{unsatisfied clause } C$ and $p_\text{walk},r_\row$
\For{literal in $C$}
\State $\text{score} \gets \text{breakcount} + r_\row \cdot (\text{\acrshort{tlc} change})$
\EndFor
\State $p\gets \Call{RandomUniform}{0,1}$
\If{all $\text{scores} > 0$ and $p < p_\text{walk}$}
\State $l\gets \text{random variable in }C$
\Else
\State $l\gets \text{variable in } C \text{ with minimal score}$
\EndIf
\State \Return $l$
\end{algorithmic}
\end{algorithm}
By viewing true literals as a finite resource, we can gain an intuitive understanding of the mechanism behind \gls{doc}.
Consider having to choose between two states with identical (and low) energy $E$, but different \gls{tlc}.
Trying to fix the remaining broken clauses demands an increase in \gls{tlc} by introducing new true literals, and because the \gls{tlc} can not be increased indefinitely, the state with fewer true literals can be interpreted as the more efficient assignment.
Additionally, from a probabilistic standpoint the chances of introducing new true literals during the process of satisfying the remaining clauses is higher in that state: the lower the \gls{tlc} is, the more likely a flip of a false literal is to produce true literals elsewhere.
As such, the \gls{tlc} amounts to a refinement of the search criterion for states beyond their energy.

Tying this back into our discussion at the end of \autoref{sec:tlc_stats}, we expect \gls{doc} to be particularly useful when the solution states are fragile in the sense of containing many critically satisfied constraints, as appears to be common for the hardest Weigt-type instances.

\section{Performance}\label{sec:experiments}
In \walksat{}, it is known empirically that $p_\text{walk}\approx 0.57$ is optimal for large random 3-SAT instances \cite{solvers.walksat.optimal_noise.Kroc_2010,SLS.ProbSAT.Prob_distributions.Balint_2012}.
To demonstrate that fine-tuning this parameter does not significantly influence the performance, we also show variants with $p_\text{walk} = 0.4$ -- which proves superior in some smaller examples -- and $p_\text{walk} = 0.5$.
For \rowsat{}, we choose $p_\text{walk} = 0.4$ and $r_\row = 0.15$.

\begin{figure}[htp!]
    \centering
    \includegraphics{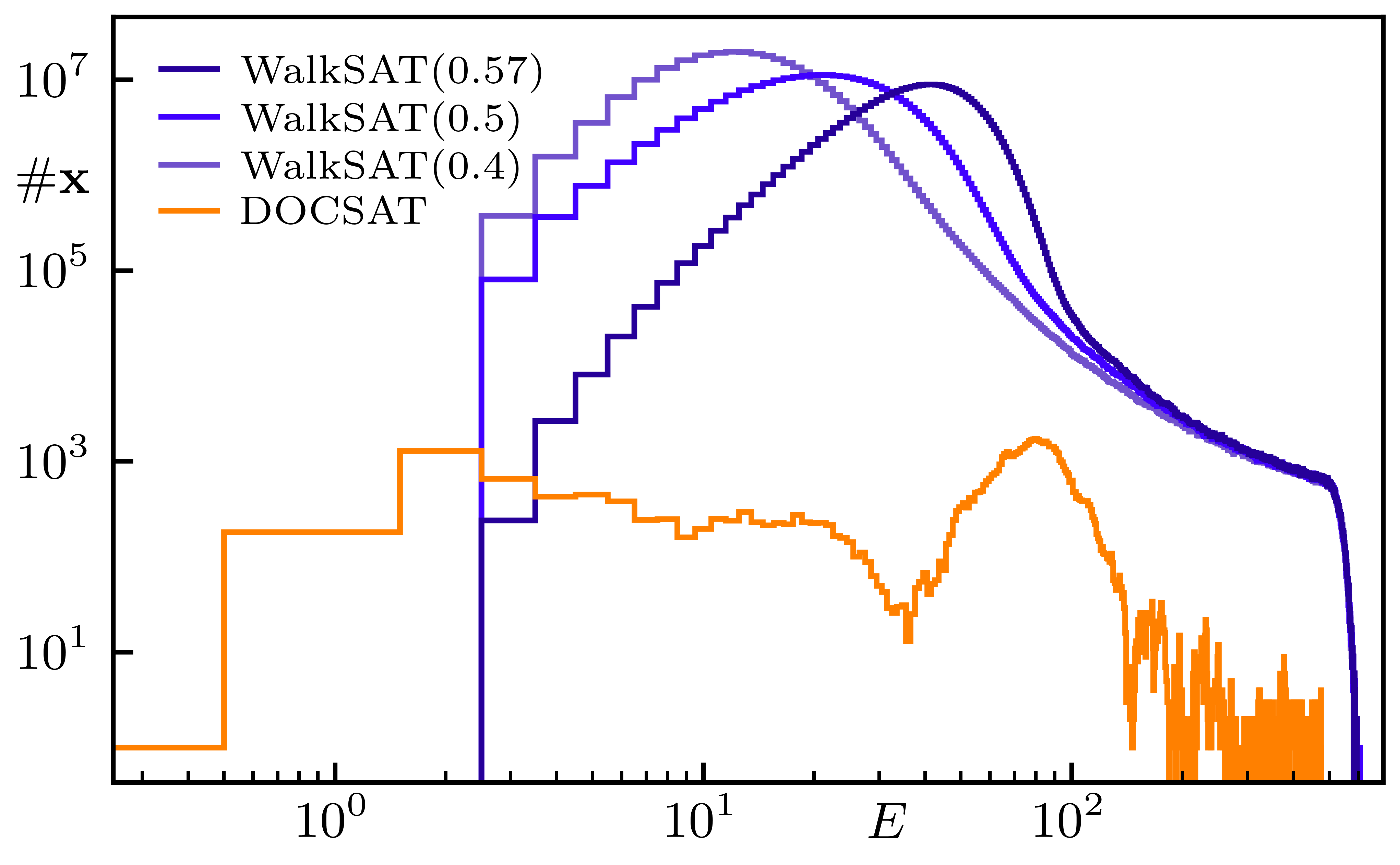}
    \caption{Logarithmic plot of the number of states $\#\textbf{x}$ with energy $E$ for a hard 3-SAT instance with $N=1000$ at $\alpha=\alpha_\text{crit}$ over $10^3$ trials with $3\cdot 10^5$ flips each.
    The walk probabilities $p_\text{walk}$ for \walksat{} are displayed in the legend; \rowsat{} uses $p_\text{walk}=0.4$ and $r_\row=0.15$.
    Note that the solution ($E=0$) is at the left of the plot and that \rowsat{} terminates after finding it, thus not exhausting all iterations.}
    \label{fig:hists_rowsat_wsat_f1000v0}
\end{figure}

With the new heuristic, we are able to solve all of the $N=200$ instances from \autoref{sec:tlc_stats}. 
The potential of the algorithm is further demonstrated in \autoref{fig:hists_rowsat_wsat_f1000v0}, where we visualize the states found by \walksat{} and \rowsat{} for a hard instance with $N=1000$.
There, we show the number of states $\#\textbf{x}$ found as a function of energy $E$ over $10^3$ trials with $3\cdot 10^5$ flips each; \rowsat{} is terminated after finding the solution, as it would just oscillate between $E=0$ and $E\ge 1$ afterward.
Note that \walksat{} gets stuck at $E=3$, and that this instance also appears out of reach for current state-of-the-art complete solvers such as Kissat \cite{solvers.kissat_et_al.SAT_comp2024.Biere_2024,solvers.cadical2.Biere_2024}. 
More generally, we stress that for the hard critical instances studied in our present work, more complex algorithms do not (significantly) outperform \walksat{}, which, apart from simplicity, motivates our focus on benchmarking our findings against \walksat{} in the following. 
In \appref{sec:app:heuristics_gallery} we provide further benchmark data for other algorithms and variants.

For a more in-depth benchmark, we analyze samples of 250 instances each for problem sizes from $N=100$ to $N=15000$ at the critical clause density $\alpha_\text{crit}=4.27$. 
For \walksat{} and \rowsat{} we run $10^3$ trials with $300N$ flips per trial, where the linear scaling in system size is to avoid an artificial decrease in success probability.
In \autoref{fig:sol_ratio_wsat} we visualize the solution ratio $R_\text{sol} = \frac{\text{solved instances}}{\text{total instances}}$ as a function of $N$, demonstrating the reliably strong performance of \rowsat{} in the critically hard realm for unprecedented system sizes.

\begin{figure}[!htp]
    \centering
    \includegraphics{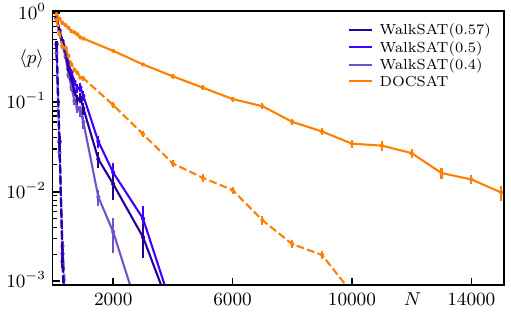}
    \caption{Average success probability $\langle p\rangle$ for the same dataset as in \autoref{fig:sol_ratio_wsat}.
    The walk probabilities $p_\text{walk}$ for \walksat{} are displayed in the legend; \rowsat{} uses $p_\text{walk}=0.4$ and $r_\row=0.15$.
    For the dashed lines, the average is taken over the hardest quintile (50 instances) for each solver.
    Errors correspond to the standard deviation of the mean.}
    \label{fig:p_avg_wsat}
\end{figure}

As a different representation of the benchmark dataset, we also show the ensemble-averaged success probability per run $\langle p \rangle$ in \autoref{fig:p_avg_wsat}, where the expected runtime for finding a solution scales as $1/\langle p \rangle$.
Additionally, the figure contains the average over the hardest quintile of instances for each solver, which filters out (comparably) easy problems.

The data is consistent with an exponential fit 
\begin{align}
\langle p \rangle \sim (1+b)^{-N}, \label{eqn:scaling}
\end{align}
in agreement with an asymptotically exponential runtime.
Quite remarkably, the asymptotic scaling of \rowsat{} is found to be substantially superior to \walksat{} in all aspects. 
Specifically, in \autoref{eqn:scaling} for the entire benchmark sample, we find $b  \approx 2\cdot 10^{-3}$ for the best-scaling \walksat{} parameter $p_{\text{walk}} = 0.57$, while $b  \approx 3\cdot 10^{-4}$ for \rowsat{}. For the hardest quintile, we instead find $b \approx 3\cdot 10^{-2}$ for \walksat{}, and $b \approx 6\cdot 10^{-4}$ for \rowsat{}, respectively. 
The latter observation does not only demonstrate that \rowsat{} performs stronger on the hardest quintile than \walksat{} does on average, but also that the relative loss of performance when comparing the entire sample with the hardest quintile is much smaller for \rowsat{}.

\begin{figure}[!htp]
    \centering
    \includegraphics{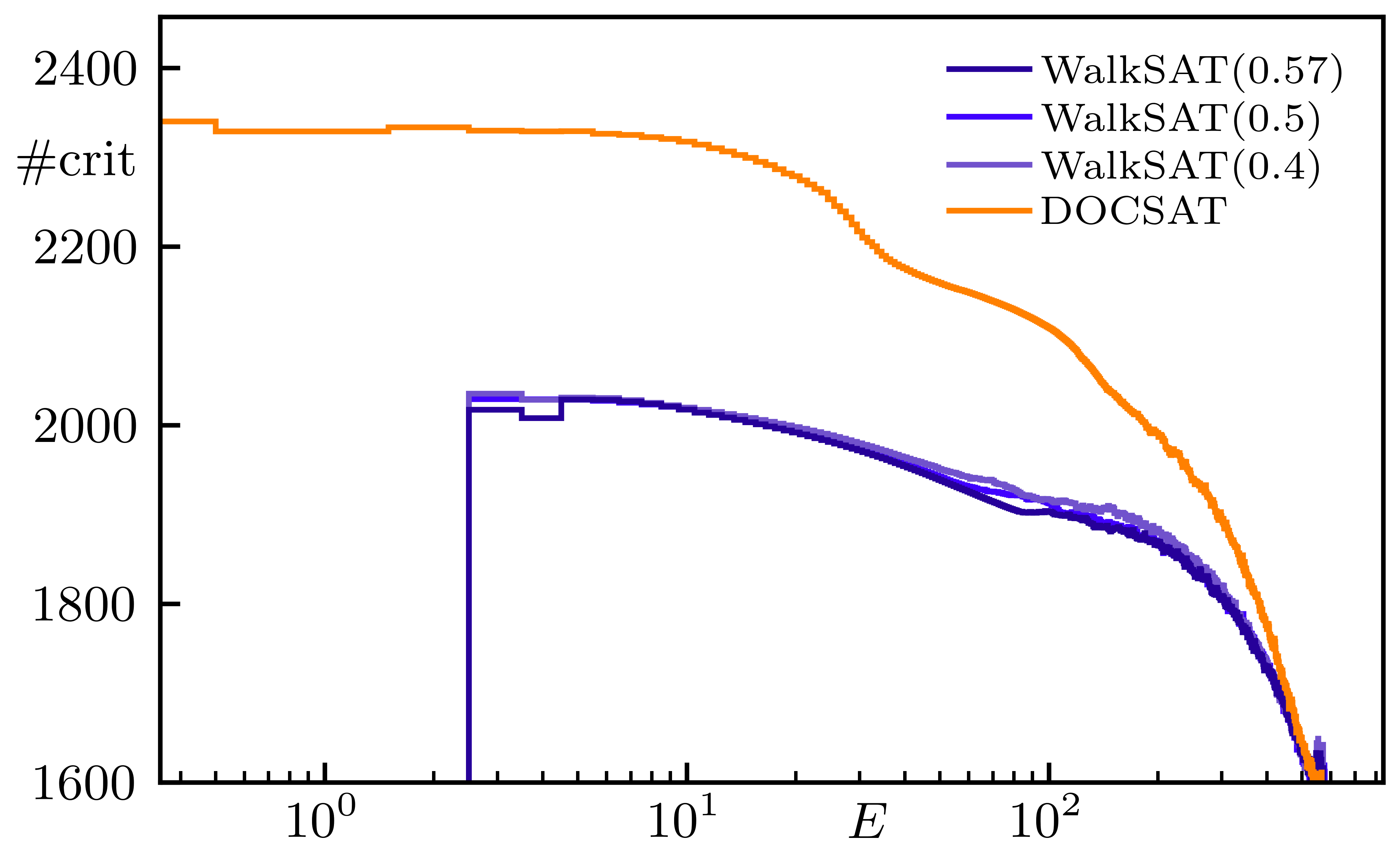}
    \caption{Average number of critical clauses {\#}crit in states with energy $E$ for the hard $N=1000$ instance from \autoref{fig:hists_rowsat_wsat_f1000v0} over $100$ trials with $3\cdot 10^5$ flips each.
    The walk probabilities $p_\text{walk}$ for \walksat{} are displayed in the legend; \rowsat{} uses $p_\text{walk}=0.4$ and $r_\row=0.15$.}
    \label{fig:critical_clauses_f1000}
\end{figure}

Finally, we find it interesting to analyze the generation of critical clauses at the expense of oversatisfied ones in \rowsat{}, which was the main intuition behind its construction (cf.~\autoref{sec:tlc_stats} and \autoref{sec:rowsat}). 
Upon selecting and flipping a variable in an unsatisfied clause, all the other clauses in which the variable appears change too.
Indeed, the rate $\Gamma_c$ at which critical clauses are increased by the non-random step (see line 8 in \algoref{alg:rowsat}) for \rowsat{} is about a factor of $4$ higher than for \walksat{} ($r_\text{doc}=0$ in \algoref{alg:rowsat}).
Also including the transitions from unsatisfied (in addition to those from oversatisfied) to critical clauses, the overall generation of critical clauses in \rowsat{} remains larger by a factor of roughly $2$. 
To illustrate this characteristic behavior in greater depth, in \autoref{fig:critical_clauses_f1000} we compare \walksat{} and \rowsat{} regarding the average number of critical clauses in configurations occurring during the (attempted) solution of a hard 3-SAT instance of size $N=1000$ (cf.~\autoref{fig:hists_rowsat_wsat_f1000v0}) as a function of energy. 
Clearly, \rowsat{} maintains a higher number of critical clauses throughout its search, and the difference to \walksat{} increases with decreasing energy, which we deem crucial to \rowsat{}'s capability of actually finding the global minimum ($E=0$).

These properties establish a specific relation between the guiding principle behind the construction of the \rowsat{} heuristic and the empirical observation of its practical performance.

\section{Concluding discussion}\label{sec:outro}
We have introduced a \acrfull{sls} heuristic coined \rowsat{} that significantly improves on existing stochastic 3-SAT solvers in the realm of particularly hard satisfiable instances by reducing the number of oversatisfied combinatorial constraints so as to promote critical clauses.
Our construction is based on the analysis of states found by \walksat{} in the low-energy regime for satisfiable 3-SAT problems at the notoriously difficult critical value of the clause density, specifically on our observation that solutions typically have fewer true literals than readily found local minima.
As our numerical simulations demonstrate, exploiting this statistical property yields substantial improvement over both \walksat{} and other modern solvers.

Adopting a broader perspective, the idea of \acrfull{doc} can be interpreted as identifying a criterion to guide the \gls{sls} toward more favorable states that is complementary to the number of violations (energy), i.e. the primary cost function of 3-SAT. 
We note that for natural physical problems, additional structure such as symmetries and spatially local correlations commonly distinguish beyond the physical notion of energy ground states from say random states.  
In this sense, inspired by the search for such additional structure, our implementation of \gls{doc} provides a minimal working example for escaping a vast landscape of local minima characterized by a certain statistical property distinguishing true solutions in the context of abstract combinatorial problems. 
In particular, moving beyond true literal counts, one could also generalize this concept to distributions of true literal patterns or directly target already satisfied constraints. 

\begin{acknowledgments}
We acknowledge discussions with Yumin Hu and Tim Pokart as well as financial support from the German Research Foundation (DFG) through the Collaborative Research Centre (SFB 1143, project ID 247310070) and the Cluster of Excellence ct.qmat (EXC 2147, project ID 390858490).
\end{acknowledgments}

\appendix

\section{Performance of other algorithms}\label{sec:app:heuristics_gallery}
\begin{figure}[!htp]
    \centering
    \includegraphics{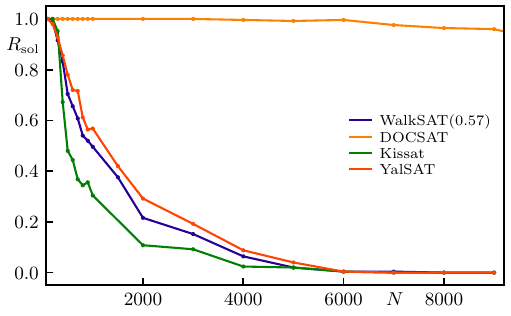}
    \caption{Fraction $R_\text{sol}$ of solved instances for algorithms \yalsat{} and \kissat{} for the same benchmark instances as in \autoref{fig:sol_ratio_wsat}.
    For comparison, we also repeat the best scaling \walksat{} with $p_\text{walk}=0.57$ and \rowsat{} with $p_\text{walk}=0.4$ and $r_\row=0.15$.}
    \label{fig:sol_ratio_kissat_demo}
\end{figure}

In this section, we illustrate the performance of two other SAT solvers, namely the \gls{sls} solver \yalsat{} (winner of the uniform-random track in 2017) \cite{sat.competition_2017.yalsat.Biere_2017}, which builds on \probsat{} \cite{SLS.ProbSAT.Prob_distributions.Balint_2012}, and the complete solver \kissat{} (winner in all main tracks in 2024) \cite{solvers.kissat_et_al.SAT_comp2024.Biere_2024,solvers.cadical2.Biere_2024}. 
Even at critical clause density, problems with $N=200$ are still sufficiently easy for \kissat{} (\yalsat{} solves all but a few of the instances that are hardest for \walksat{}, cf.~\autoref{sec:tlc_stats}). 
However, although one can often solve practical problems with millions of variables in many applications, critically hard random instances -- such as the Weigt-protocol instances studied in our present benchmark -- already become challenging at much smaller $N$ even for these most advanced solvers (see \autoref{fig:sol_ratio_kissat_demo}).

Due to the significantly more complex nature of these solvers, we refer to the respective original publications for a detailed description.
Both solvers are run with the default settings used for the competitions (except hinting \kissat{} at targeting satisfiable instances).
\yalsat{} follows the usual \gls{sls} structure of using heuristics to select clauses and variables, so we set the same limit of $300N$ flips per trial as for \walksat{} and \rowsat{}.
Because the heuristics are more involved, this results in a slightly longer runtime granted to \yalsat{}.
For \kissat{}, we instead use a cutoff on the maximal number of so-called decisions during the search. While this has the advantage of being a hardware-independent measure, we note that the runtime scales slightly superlinear in the decisions.
Since the solver is complete we only run it once with at least comparable total walltime to compensate for the $10^3$ restarts of the \gls{sls} solvers. 
Specifically, using $3000N$ decisions in \kissat{} yields similar walltime as \yalsat{} at $N=500$ with an additional factor of up to 2 in favor of \kissat{} toward $N=9000$.
In \autoref{fig:sol_ratio_kissat_demo}, we compare the fraction of solved instances for all aforementioned solvers with \rowsat{}, further corroborating its superior performance  (cf.~\autoref{fig:sol_ratio_wsat}).
\,\\
\begin{figure}[H]
    \centering
    \includegraphics{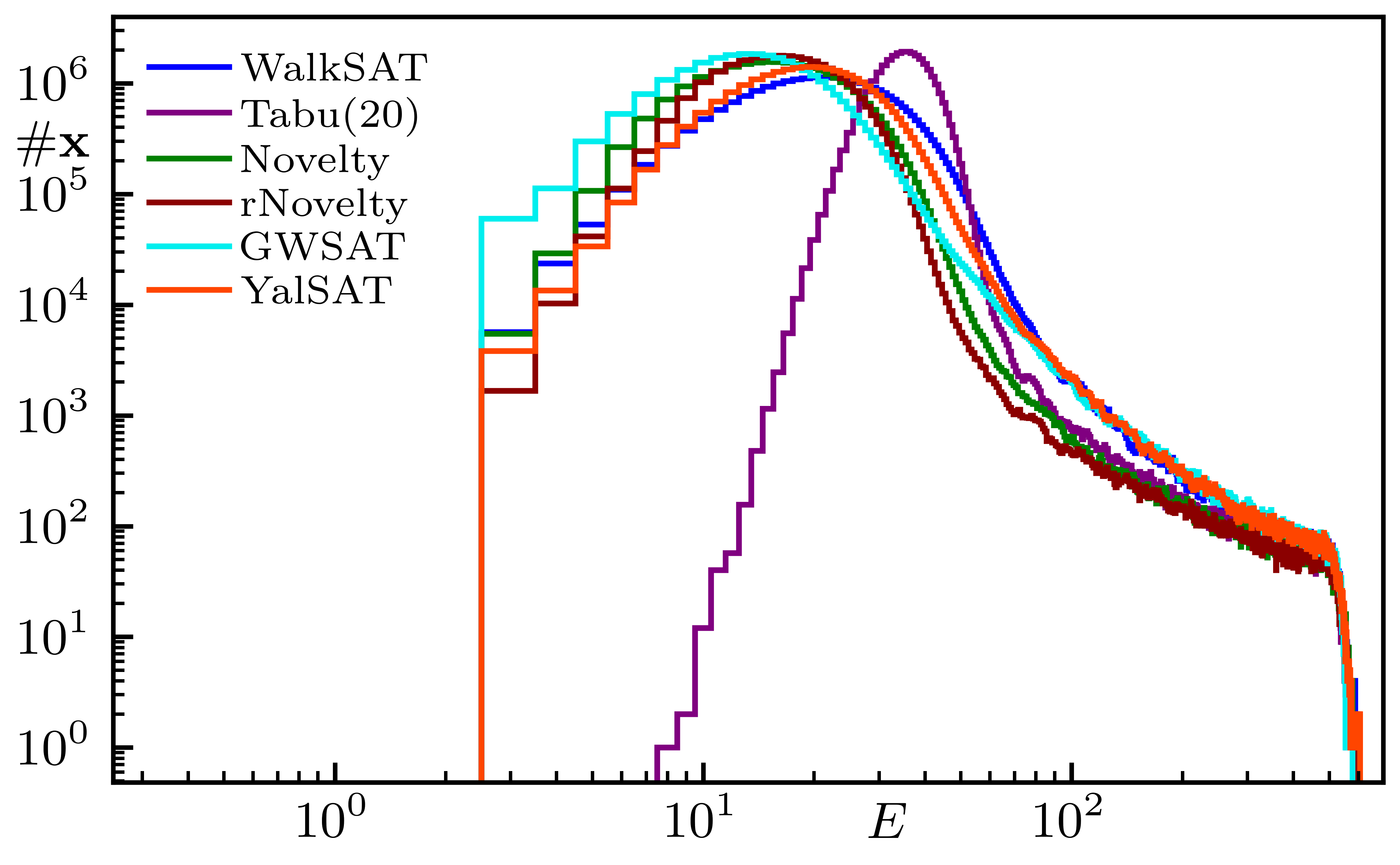}
    \caption{\walksat{} and related \gls{sls} algorithms for a hard $N=1000$ instance readily solved by \rowsat{} (cf.~\autoref{fig:hists_rowsat_wsat_f1000v0}).
    The walk probability is fixed to $p_\text{walk}=0.5$; see the main text for further information on the displayed solvers and references.}
    \label{fig:wsat_ysat_f1000_demo}
\end{figure}

In \autoref{fig:wsat_ysat_f1000_demo} we benchmark \yalsat{} on the hard $N=1000$ instance used for \autoref{fig:hists_rowsat_wsat_f1000v0}.
There we also include some of the (relatively minor) variations on \walksat{} such as Novelty and GWSAT \cite{SLS.adaptive_noise.novelty+.G2WSAT.Li_2007,SLS.gNovelty+.Pham_2007}.
Tabu is \walksat{} with a list of recently flipped variables (here 20) that are not allowed to be flipped back; this avoids loops, but evidently is not very helpful for these random instances.
GWSAT takes into account how many unsatisfied clauses are fixed by flipping a variable (so-called makecount $m$) and optimizes $b-m$ instead of just $b$ (corresponds to the greedy algorithm GSAT combined with random walks).
Curiously, even \yalsat{} follows a very similar curve, while the behavior of \rowsat{} is drastically different and converges to the full solution well before exhausting the iteration limit (cf.~\autoref{fig:hists_rowsat_wsat_f1000v0}).
\FloatBarrier
\bibliography{bibliography}

\end{document}